\documentclass[lettersize,journal]{IEEEtran}
\usepackage{amsmath,amsfonts}
\usepackage{algorithmic}
\usepackage{algorithm}
\usepackage{array}
\usepackage{textcomp}
\usepackage{stfloats}
\usepackage{url}
\usepackage{verbatim}
\usepackage{cite}
\usepackage{multirow}
\usepackage{enumerate}
\usepackage[utf8]{inputenc}
\usepackage[T1]{fontenc}
\usepackage{hyperref}
\usepackage{url}
\usepackage{booktabs}
\usepackage{amsfonts}
\usepackage{nicefrac}
\usepackage{microtype}
\usepackage{xcolor}
\usepackage{amsmath}
\usepackage{amsthm}
\usepackage{graphicx}
\usepackage[justification=centering]{caption}
\usepackage{subfigure}

\begin{document}

\title{Adaptive Sparse Softmax: An Effective and \\Efficient Softmax Variant}

\author{Qi Lv$^\dag$, Lei Geng$^{\dag}$, Ziqiang Cao, Min Cao, Sujian Li, Wenjie Li, and Guohong Fu
    \thanks{$^{\dag}$means the equal contribution.}
    \thanks{Qi Lv is with the School of Computer Science and Technology, Soochow University and the School of Computer Science and Technology, Harbin Institute of Technology (Shenzhen) (e-mail: aopolin.ii@gmail.com).}
    \thanks{Lei Geng is with the School of Computer Science and Technology, Soochow University (20215227001@stu.suda.edu.cn).}
    \thanks{Ziqiang Cao, Min Cao, Guohong Fu are with the School of Computer Science and Technology, Institution of Artificial Intelligence, Soochow University. (Corresponding author: Ziqiang Cao, zqcao@suda.edu.cn)}
    \thanks{Sujian Li is with the Peking University and Wenjie Li is with the Hong Kong Polytechnic University.}
}

\markboth{IEEE/ACM TRANSACTIONS ON AUDIO, SPEECH, AND LANGUAGE PROCESSING,
Vol. 1, No. 1, January 2021}
{Lei Geng, Qi Lv,
\MakeLowercase{\textit{(et al.)}:
Adaptive Sparse Softmax: An Effective and Efficient
Softmax Variant.}}

\maketitle

\begin{abstract}
Softmax with the cross entropy loss is the standard configuration for current neural classification models.
The gold score for a target class is supposed to be 1, but it is never reachable under the softmax schema.
Such a problem makes the training process continue forever and leads to overfitting.
Moreover, the ``target-approach-1'' training goal forces the model to continuously learn all samples, leading to a waste of time in handling some samples which have already been classified correctly with high confidence, while the test goal simply requires the target class of each sample to hold the maximum score.
To solve the above weaknesses, we propose the \textbf{A}daptive \textbf{S}parse softmax (AS-Softmax) which designs a reasonable and test-matching transformation on top of softmax.
For more purposeful learning, we discard the classes with far smaller scores compared with the actual class during training.
Then the model could focus on learning to distinguish the target class from its strong opponents, which is also the great challenge in test.
In addition, since the training losses of easy samples will gradually drop to 0 in AS-Softmax, we develop an adaptive gradient accumulation strategy based on the masked sample ratio to speed up training.
We verify the proposed AS-Softmax on a variety of text multi-class, text multi-label, text token classification, image classification and audio classification tasks with class sizes ranging from 5 to 5000+.
The results show that AS-Softmax consistently outperforms softmax and its variants, and the loss of AS-Softmax is remarkably correlated with classification performance in validation.
Furthermore, adaptive gradient accumulation strategy can bring about 1.2$\times$ training speedup comparing with the standard softmax while maintaining classification effectiveness.
\end{abstract}


\section{Introduction}\label{sec_intro}
\IEEEPARstart{S}{oftmax} is widely used as the last-layer activation function of a neural classification model.
It normalizes the output of a network to a probability distribution over predicted output classes.
Softmax is extremely appealing in both academics and industries as it is simple to evaluate and differentiate.
Many research efforts are devoted to refining softmax.
On one hand, several works~\cite{1, 2} have explored accelerating the training process by outputting a subset of classes to reduce computational complexity.
On the other hand, improving training effectiveness is also an important direction.
For example,~\cite{3} and~\cite{4} enlarged the margin of intra-class and inter-class in softmax, while~\cite{5} used label smoothing to prevent the neural networks from excessively trusting gold labels.

Despite the probability form, each output score of softmax is within the range of $(0,1)$, unlike the sparse probability distribution in real classification tasks.
\textcolor{black}{
Consequently, Softmax requires continuous optimization towards probability 1 for the target class, even when the model has already learned to rank it correctly with a sufficient margin. 
This unnecessary optimization can lead to overfitting by forcing the model to fit the training data more strongly than needed for correct classification~\cite{6}.
}
\textcolor{black}{
In addition, in classification tasks, Softmax with cross-entropy loss optimizes the probability of the target class toward 1, which may not be necessary for correct classification that only requires maintaining correct relative order among classes.}
Thus there is an obvious gap between the two objectives.
Take the following two prediction cases as an example:
It is a 5-class classification task and Class 1 is the gold label.
\begin{description}
    \item[Case A:] \ \ \{\textbf{0.4},\ 0.15,\ 0.15,\ 0.15,\ 0.15\}
    \item[Case B:] \ \ \{\textbf{0.49},\ 0.5,\ 0.004,\ 0.003,\ 0.003\}
\end{description}
We can find that in Case A, the score of Class 1 is far larger than all the other classes, leading a successful prediction.
By contrast, there is a strong negative class in Case B and its classification result is wrong.
Unfortunately, in the training period, since the target score in Case B is much higher than that in Case A,
\textcolor{black}{the cross-entropy loss for case A is 0.92, and for case B, it is 0.71, implying that the gradient update direction of model parameters is more influenced by the former.
However, evidently, the prediction of case A is correct while case B's prediction is incorrect. The optimization goal for the model should be to make both predictions correct. In other words, we believe that the loss generated by case B should have a greater guiding effect on the model parameter updates.}
Our experiments show that sometimes the correlation coefficient between the cross entropy loss of softmax and the classification accuracy is even close to 0 in validation (refer to Table~\ref{tab:pearsonr results}).

To address the above-mentioned deficiency, we propose the \textbf{A}daptive \textbf{S}pare softmax (AS-Softmax) which involves a reasonable and test-matching transformation on top of softmax.
Specifically, when computing the softmax activation function, we exclude the classes whose scores are smaller than the actual class to a given margin.
In this way, the model focuses on learning to distinguish the target class from its strong opponents, matching the use in test.
It is also noted that the loss of AS-Softmax can drop to 0 as long as the score of the target class is much larger than all other classes.
In other words, with the progress of training, more and more easy samples will be dropped from back propagation.
Thus the learning process tends to find the useful hard training instances, which makes learning more effective.
In practice, it is not always the case that increasing the training sample helps the final performance of the model.
\cite{7} and~\cite{8} claimed that training the model with representative instances is more efficient.
Easy samples can be discarded by the cascade structure at early stages~\cite{9} for learning better classification models.
Moreover, the increasing removed samples also guide us in developing a novel adaptive gradient accumulation strategy to make the training process faster and faster.

We test AS-Softmax on a variety of classification tasks, including text multi-class classification~\cite{10,11}, text multi-label classification~\cite{12,13}, token classification~\cite{14, 15}, image classification~\cite{56} and audio classification~\cite{57}.
All the benchmarks are publicly available and the class sizes range from 5 to 5000+.
Experiments show that AS-Softmax consistently improves the performance of text classification and alleviates overfitting to a large extent.
In addition, AS-Softmax with the adaptive gradient accumulation strategy can bring about 1.2$\times$ training speedup while maintaining the performance.

To conclude, AS-Softmax solves the endless training and train-test objective mismatching problems of the original softmax function. 
Its advantages can be summarized as follows:
\begin{itemize}
    \item With AS-Softmax, easy samples will be dropped from back propagation gradually, moderating overfitting and making training more effective.
    \item The loss of AS-Softmax is highly correlated with the final classification performance.
    \item Its training can be accelerated via the proposed adaptive gradient accumulation strategy.
    \item AS-Softmax is fairly easy to implement. We have made it as an standard module in Pytorch\footnote{https://github.com/aopolin-lv/AS-Softmax}.
\end{itemize}

\section{Related Work}
Softmax loss function is widely used as an output activation function for modeling categorical probability distributions.
There are a lot of variants of softmax proposed in the literature~\cite{16, 17, 18, 19}.
We will briefly introduce them in terms of efficiency and effectiveness.
\subsection{Enhancing Efficiency}
Training a model using the full softmax loss becomes prohibitively expensive in the settings where a large number of classes are involved.
One important kind of efficient training is to reduce its output dimension to reduce computational complexity.
For example, hierarchical softmax~\cite{20} partitioned the classes into a tree based on class similarities.
Unlike hierarchical softmax, Differentiated softmax (D-Softmax)~\cite{48} could speed up the training and inference process. 
\cite{48} thought that high-frequency words should be fitted with more parameters, while low-frequency words should be fitted with fewer parameters. 
Meanwhile, many works explored efficient subsets instead of all output classes.
We define these methods as \textit{sparse softmax family} uniformly.
\cite{21} proposed an importance sampling training algorithm which could effectively keep the computational complexity during training as using small vocabulary for neural machine translation.
\cite{22} introduced a new kernel-based sampling
method RF-softmax which employs the powerful Random
Fourier Features and guarantees small bias of the gradient estimate.
\cite{23} proposed a sampling distributions that are adaptive to the model’s input, structure, and the current model parameters.
\cite{17, 24} suggested a new softmax variant which could generate sparse posterior distributions.

Some studies also tried to use approximate estimation to simplify the calculation of softmax.
\cite{25, 26} used self-normalization mechanism and noise contrastive estimation separately to simplified the complexity of calculation, respectively.
Spherical softmax~\cite{27, 16} adopted a quadratic function instead of the original the exponential function, making its parameter update independent of the output size while~\cite{1} introduced the singular value decomposition method to restrict the probability distribution.
\cite{50, 51, 52, 53} utilized approximate nearest neighbor method to select training class to approximate full training data.
Others~\cite{52, 53} attempted to accelerate the softmax computation from hardware perspective.
It allows to simplify the complexity of hardware and signal propagation delay, splitting the exponentiation
calculation of the softmax into several specific basics.

\subsection{Boosting Effectiveness}
Softmax usually lacks the accurate discrimination of similar output classes.
\cite{3} found the result distribution learnt from softmax activation function is distributed radially. 
Thus, an intuitive idea is to enlarge the inter-class margin and compress the intra-class distribution.
\cite{4} designed a soft distant margin that only modified the forward of softmax without changing the backward and could be easily optimized by the typical optimizers while~\cite{3} incorporated angular margin to obtain a more discriminative decision boundary.
The features learned by softmax loss have intrinsic angular distribution~\cite{2}.
Follow-up studies~\cite{28, 29, 30, 31, 2, 32} made more effort on the angular constraints and normalization.

Additionally, variants in the sparse softmax family like~\cite{33, 6} are also in favor of improving the effectiveness of models as they reduce the risk of overfitting in high-dimensional classification problem.
To prevent the network from becoming over-confident in the current labels,~\cite{5} introduced label smoothing algorithm that give a probability to each non-target class so that the model could learn all classes' features.
\cite{34} proposed Noise-Aware-with-Filter to distinguish hard samples from noisy labels.
Moreover,~\cite{35} overlay binary masking variables over class output probabilities by dropout to input-adaptively learned via variational inference.

Although many methods have been proposed to solve the problem of time or efficiency, the algorithm proposed in our paper is not only easy to understand and effective but also maintains good performance by automatically discarding easy samples for acceleration.
Thus we propose the \textbf{A}daptive \textbf{S}parse softmax (AS-Softmax).

The most closely related works to our method include sparsemax and sparse-softmax.
Sparsemax sparsifies the probability matrix of the results, but the range of sparsity threshold is infinite. 
Thus, it does not facilitate finding the optimal solution through a convenient method like grid search.
Sparse-softmax samples the top-k largest classes and computes their losses, but it cannot decrease the loss of simple samples to zero.
However, our approach is based on the practical needs in the usage of classification tasks: ensuring that the correct class score is higher than others, automatically converting the loss of simple samples to zero. In experiments, it performs well, operates at high speed, and significantly reduces overfitting. 

Although various softmax variants have been proposed, AS-Softmax differs from them in several key aspects. 
Unlike Sparse-Softmax~\cite{16} which retains a fixed top-k classes during training, AS-Softmax adaptively determines which samples to preserve based on their difficulty through a margin criterion $\delta$. 
Compared to Sparsemax~\cite{17} and Entmax~\cite{24} that achieve sparsity through geometric projection onto the probability simplex, AS-Softmax takes a more intuitive approach by directly matching the training objective with testing goal. 
Our margin-based criterion leads to a highly interpretable sparsification process - samples are masked when the target class probability exceeds other class probabilities by the margin $\delta$.

More importantly, while methods like AM-Softmax~\cite{28, 29, 30, 31, 2, 32} focus on modifying loss functions to improve model effectiveness, AS-Softmax uniquely combines effectiveness and efficiency through its adaptive gradient accumulation strategy. This allows the model to dynamically adjust the training process based on the ratio of masked samples, leading to significant speedup without compromising performance. Additionally, unlike existing methods where the correlation between loss and accuracy can be weak (e.g., in AM-Softmax), the loss of AS-Softmax demonstrates strong correlation with classification performance, providing a more reliable training signal.

\section{Methods} \label{sec:method}
\subsection{Background of Softmax}
Suppose the final output of the neural classifier is $o \in \mathbb{R}^{n}$.
Here ${n}$ denotes the number of classes.
The prediction score of a class $p_i$ in the original softmax is computed as follows:

\begin{align}\label{eq:softmax}
    p_i = \operatorname{softmax}({o_i}) = \frac{e^{o_{i}}}{\sum_{j=1}^{n}e^{o_{j}}}. 
\end{align}

Classifiers usually adopt cross entropy combined with softmax as their loss function.
For multi-class classification, there is only one actual target class $t$.
Then the expression of the loss function and its backward propagation are as follows:

\begin{align}
    \mathcal{L} &= - \log(p_t) \\ &= \log(\sum_{j=1}^{n}e^{o_{j}})-o_t,\\
    \nabla{\mathcal{L}} &= \frac{\partial{\mathcal{L}}}{\partial{o_j}} = 
                    \begin{cases}
                        p_j-1, &\text{if }j=t, 
                        \\p_j,&\text{otherwise.}
                    \end{cases}
\end{align}

The cross entropy loss is composed of exponential and logarithmic functions, which is convenient for computation of the loss function and its back propagation.
However, as mentioned in Section~\ref{sec_intro}, softmax is trapped in the endless training and train-test goal mismatching problems.
In addition, the smaller the cross entropy loss, the stricter the constraints between the target and non-target classes.
\cite{6} proved that in order to make sure the loss $\mathcal{L}$ can be reduced to $\log{2}$, the output should satisfy the following inequality:
\begin{align}
    o_{t}-o_{min} \ge \log(n-1),
    \label{eq:softmax requirement}
\end{align}
where $o_{t}$, $o_{min}$ respectively means the target (maximal) logit and the minimal logit.
When dealing with the high-dimensional classification problems, $\log{(n - 1)}$ is a relatively massive but unnecessary margin, which increases the overfitting risk.

In order to mitigate disadvantages aforementioned, we propose an alternative variant of softmax named \textbf{A}daptive \textbf{S}parse softmax (AS-Softmax).

\subsection{Adaptive Sparse Softmax} \label{sec:AS-Softmax}
The training goal of softmax is to make the score of the target class as large as possible.
This practice does not match the test objective to encourage the target probability greater than that of any other category.
Motivated by Sparse-softmax~\cite{6}, we propose a new training objective which encourages the probability of target class $p_t$ to exceed the probability of non-target class $p_{i\neq t}$ by a specific margin $\delta$:
\begin{align}
    p_t - p_{i\neq t} \geq \delta, \label{eq:criterion}
\end{align}
where $\delta \in \textcolor{black}{(}0,1]$ is a hyper-parameter.

To achieve this goal, we discard the classes which have already satisfied Eq.~\ref{eq:criterion} in training. 
Specifically, we adopt a binary factor $z_i$:

\begin{equation}
    z_i =
        \begin{cases}
            0,  & \text{if Eq.~\ref{eq:criterion} is satisfied and $i \neq t$,} \\
            1, & \text{otherwise.} \\ 
        \end{cases}
        \label{eq:z_i}
\end{equation}

Then the modified probability of a class $\tilde{p_i}$ is computed as follows:
\begin{equation}
    \tilde{p_i} = \operatorname{AS-Softmax}(o_i) = \frac{z_{i}{e^{o_{i}}}}{\sum_{j=1}^{n}z_{j}{e^{o_{j}}}}. \label{eq:AS-Softmax}
\end{equation}
It can be seen that AS-Softmax is extremely easy to implement.
Given the output of softmax, AS-Softmax only needs a simple linear screening step while the back propagation process remains the same.
With the import of $z_i$, we find that the losses of more and more training samples will reduce to zero, that is to say, these samples are masked, as shown in Figure~\ref{fig:ignore_samples}.
Therefore, the classification model with AS-Softmax tends to discard easy samples and focus on the rest hard cases, making learning more effective.

In the mathematical forms, AS-Softmax is close to Sparse-Softmax which preserves the fixed top $k$ negative classes to speed up training.
However, the loss of Sparse-Softmax is still always larger than 0 and it does not contrast the representations of positive and negative classes.

We also examine a simple training criterion expressed as:
\begin{align}
    \log(p_t) - \log(p_{i\neq t})=o_t - o_{i\neq t} \geq \delta'.
    \label{eq:log_delta}
\end{align}
Nevertheless, $o$ can take any value, namely no upper bound of $\delta'$, which makes it hard to find the optimal value.
More comparisons of the above two strategies are provided in following section.

\subsubsection{Algorithm Discussion}
We can prove that our training object enables the output $o$ to obey the following condition:
\begin{align}
    o_{t} - o_{min} \geq \log{(n\delta + 1)}. \label{eq:our_training_goal}
\end{align}

\begin{proof}
    From our criterion in Eq.~\ref{eq:criterion}, we can get:
    \begin{align}
       p_t-p_{min}= \frac{e^{o_t}}{\sum^{n}_{j=1}e^{o_j}} - \frac{e^{o_{min}}}{\sum^{n}_{j=1}e^{o_j}} \geq \delta.
    \end{align}
    
    This formula can be further converted into:
    \begin{align}
        e^{o_t} - e^{o_{min}} \geq \delta \sum^{n}_{j=1}e^{o_j}  \geq n \delta e^{o_{min}}.
    \end{align}
    Namely:
    \begin{align}
        e^{o_t}\geq (n\delta+1) e^{o_{min}}. 
    \end{align}
    
    With a $\log$ transformation, we finally conclude that:
    \begin{align}
        o_{t} - o_{min} \geq \log{(n\delta + 1)}.
    \end{align}
\end{proof}
Recall the requirement of softmax in Eq.~\ref{eq:softmax requirement}, the criterion of AS-Softmax is much easier to meet, as $\delta$ is quite small in practice. 

\subsubsection{Extension to Multi-Label Classification} 
Softmax is typically used in multi-class classification and token classification tasks.
However,~\cite{36} demonstrated that softmax could also be applied to multi-label classification tasks with the following loss function:
\begin{align}
    \mathcal{L} &= \log(1+\sum_{i\in\Omega_{neg}}e^{o_{i}}) + \log(1+\sum_{t\in\Omega_{pos}}e^{-o_{t}}),
    \label{eq:multi-label loss}
\end{align}
where $\Omega_{pos}$ stands for the set of target classes while $\Omega_{neg}$ denotes the set of non-target classes.
The training objective of this method is to raise scores of target classes to exceed zero and decrease scores of non-target classes to be lower than zero.
Thus in test, the classes with the scores larger than 0 can be regarded as the prediction output.
Compared with the sigmoid based method, this method can mitigate the sparse problem when the number of entire output classes is greatly larger than the number of target classes.

Likewise, we extend AS-Softmax to multi-label classification, aiming to encourage probabilities of entire target classes to exceed probabilities of non-target classes by a specific margin $\delta$.
To formulate, our additional goal is:
\begin{align}
    p_t^{min} - p_{i} &\geq \delta, \label{eq:criterion_mul} \\
    p_t - p_{i}^{max} &\geq \delta, \label{eq:criterion_mul2}
\end{align}
where $p_t^{min}$ represents the smallest probability of all target classes and $p_{i}^{max}$ means the largest probability of all non-target classes.

To achieve this goal, we define the following binary factors $z_i$ and $z_t$ for multi-label classification tasks:
\begin{align}
    z_i =
        \begin{cases}
            0,  & \text{if Eq.~\ref{eq:criterion_mul} is satisfied,} \\
            1, & \text{otherwise,}  
        \end{cases}\\
    z_t =
        \begin{cases}
            0,  & \text{if Eq.~\ref{eq:criterion_mul2} is satisfied,} \\
            1, & \text{otherwise.} 
        \end{cases}
        \label{eq:z_i_multi_label2} 
\end{align}

Consequently, the final loss function combined with AS-Softmax can be obtained as follows:
\begin{align}
    \mathcal{L} &= \log(1+\sum_{i\in\Omega_{neg}}{z_i}e^{o_{i}}) + \log(1+\sum_{t\in\Omega_{pos}}{z_t}e^{-o_{t}}).
    \label{eq:multi-label AS-Softmax loss}
\end{align}

Note, if $p_t^{min} - p_{i}^{max} \geq \delta$ is satisfied, the whole loss will reduce to 0, namely the sample masked.

\subsection{Adaptive Gradient Accumulation Strategy} \label{sec:as-speed}
\begin{figure}[!t]
\centering
\includegraphics[width=0.45\textwidth, height=6.3cm]{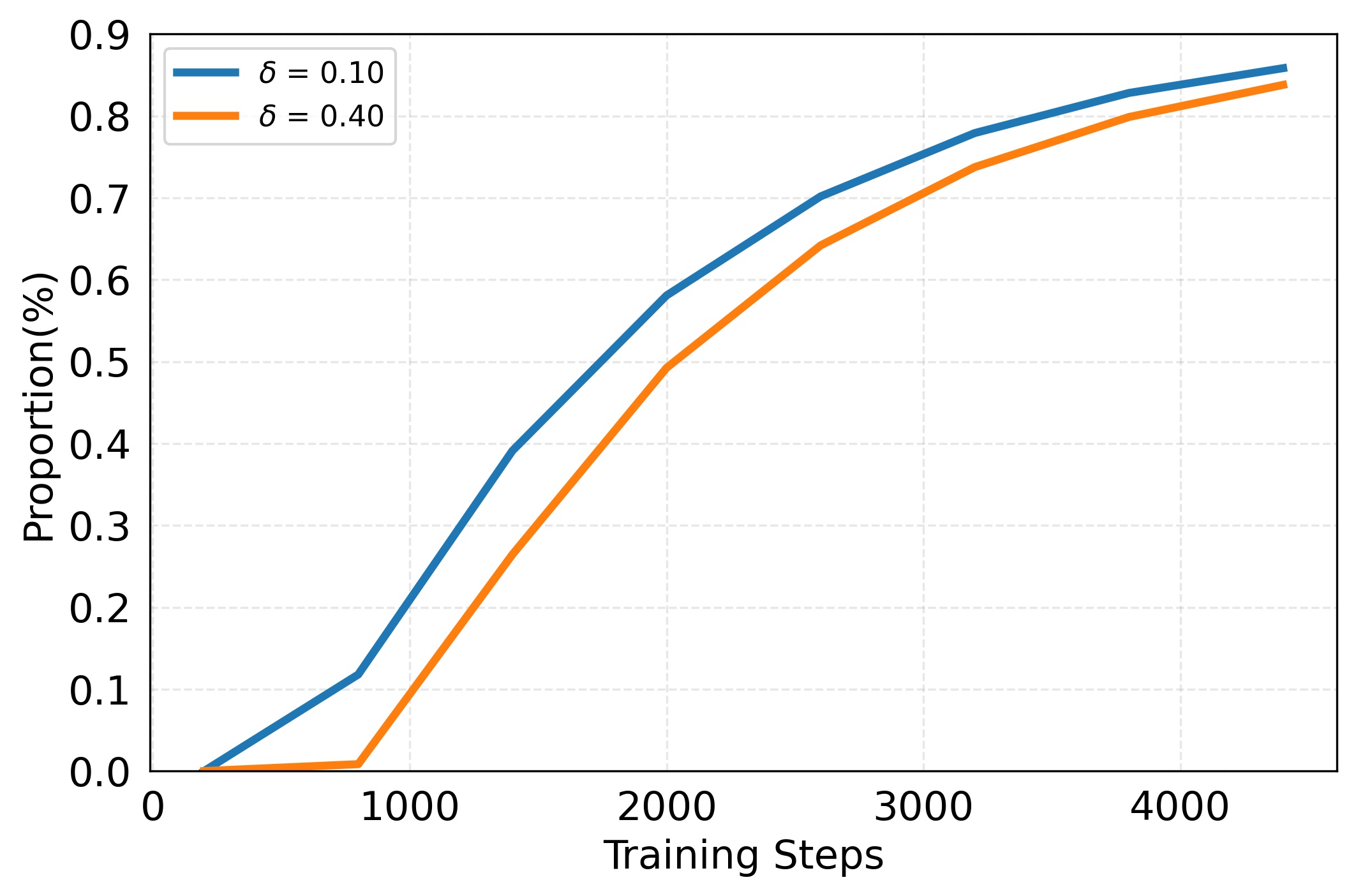}
\caption{Variation of masked samples ratio on Clinc\_oos.}
\label{fig:ignore_samples}
\end{figure}

In theory, the model with the training objective of AS-Softmax will gradually remove easy samples during training.
As shown in Figure~\ref{fig:ignore_samples}, experimental result shows that the effective sample size in a batch of AS-Softmax is much smaller than that of softmax.
Such few effective samples in a batch would waste computing resources such as the GPU memory.
In addition, the hyper-parameters of the training optimizer such as the learning rate do not well match the training process in softmax.
Therefore, we accumulate the training steps of AS-Softmax to make the effective sample size approach that of softmax.
Specifically, we propose an adaptive gradient accumulation method according to the masked samples ratio:
\begin{align}
    steps_{accum}=\lambda * \frac{N_{all}}{N_{all} - N_{masked}},
\end{align} 
where $\lambda$ is a hyper-parameter controlling the accelerate magnitude and $N_{all}$/$N_{masked}$ denotes the number of all/masked samples.
$\lambda$ is necessary as $steps_{accum}$ is a prediction of the successive batches. 

In addition, we attach the following restrictions to the accelerate strategy to guarantee its effectiveness:
i). making sure $steps_{accum}$ is increasing monotonically;
ii). ensuring the difference between two adjacent $steps_{accum}$ does not exceed 1; and
iii). setting an upper limitation to $step_{accum}$.
For convenience, we call the acceleration algorithm of AS-Softmax as AS-Speed.

\section{Experiments}\label{sec:main_result}
\subsection{Datasets and Evaluation Metrics}
To evaluate our approach extensively, we conduct experiments on 6 different text classification datasets, 1 image classification dataset and 1 audio classification dataset, including SST5~\cite{10}, Clinc\_oos~\cite{14}, Conll2003~\cite{11}, SIGHAN2015~\cite{15}, Eurlex~\cite{37}, WOS-46985~\cite{13}, WikiArt~\cite{56} and Chest\_falsetto~\cite{57}.
These datasets belong to text multi-class classification, text token classification, text multi-label classification, image classification and audio classification tasks respectively.
We separately adopt following evaluation metrics: accuracy, f1, macro/micro-f1 to measure multi-class classification, token classification and multi-label classification tasks.
\textcolor{black}{
Our criterion for selecting evaluation metrics is based on the metrics used in the leaderboard of each dataset. 
Therefore, different datasets have different evaluation metrics.
}
The basic information of the datasets and \textcolor{black}{detailed} evaluation metrics is listed in Table~\ref{tab:statistics}.

\begin{table*}[htbp]
  \centering
  \small
    \begin{tabular}{c|c|c|c|c|c}
    \toprule
    \textbf{Dataset} & \multicolumn{1}{c|}{\textbf{Task}} & \# \textbf{Train/Dev/Test} & \multicolumn{1}{l|}{\# \textbf{Classes}} & \textbf{Model} & \textbf{Metrics} \\
    \hline
    \midrule
    SST5  & \multirow{2}[0]{*}{Text multi-class classification} & 8,544/1,101/2,210 & 5     & BERT  & accuracy \\
    Clinc\_oos &       & 15,250/3,100/5,500 & 151   & BERT  & accuracy \\
    \midrule
    Conll2003 & \multirow{2}[0]{*}{Text token classification} & 14,042/3,251/3,454 & 9     & BERT  & f1 \\
    SIGHAN2015 &       & 277,786/1,100/1,100 & 5,201  & Roberta & f1 \\
    \midrule
    Eurlex & \multirow{2}[0]{*}{Text multi-label classification} & 55,000/5,000/5,000 & 127   & BERT  & macro/micro-f1 \\
    WOS-46985 &       & 30,080/7,518/9,397 & 141   & BERT  & macro/micro-f1 \\
    \midrule
\textcolor{black}{WikiArt} & \textcolor{black}{Image   classification} & \textcolor{black}{4,493/1,295/629}     & \textcolor{black}{27}         & \textcolor{black}{ViT} & \textcolor{black}{accuracy} \\
\textcolor{black}{Chest\_falsetto}               & \textcolor{black}{Audio   classification} & \textcolor{black}{1,024/128/12}       & \textcolor{black}{4}          & \textcolor{black}{Wav2vec2}            & \textcolor{black}{accuracy} \\ 
    \bottomrule
    \end{tabular}
    \caption{Statistics of experimental datasets.}
  \label{tab:statistics}
\end{table*}

\textbf{SST5.} 
This dataset~\cite{10} is a sentence level classification dataset for sentiment analysis.
It contains 5 output classes (from very negative to very positive).
It is made up of 8,544 train, 1,101 validation, 2,200 test samples.
We adopts pretrained language model BERT as the text encoder which is initialized with Bert-base-cased parameters.
The evaluation metric for this dataset is accuracy.

\textbf{Clinc\_oos.}
This dataset~\cite{11} is for intent classification, comprising of 151 classes over 10 domains. 
Its train/validation/test dataset contains 15,250/3,100/5,500 instances.
We use BERT as our backbone.
Accuracy is also used as the evaluation metric.

\textbf{Conll2003.}
This dataset~\cite{14} is for named entity recognition (NER) task which consists of four types of name entities: persons, locations, organizations and others.
There are 9 categories in its output.
The sizes of train/validation/test dataset are 14,042/3,251/3,454, respectively.
We use f1 score to measure this dataset.

\textbf{SIGHAN2015.}
This dataset~\cite{15} is designed for the Chinese Spelling Check (CSC) task.
Following previous works~\cite{38, 39}, we formula CSC as a token classification task whose candidates are tokens in vocab.
The training dataset with the size of 277,786 is comprising of the automatic generated dataset~\cite{40} and all SIGHAN training sets~\cite{41, 42, 15}.
The test set is the SIGHAN2015 test~\cite{15} with a size of 1,100.
Considering the tokens we focus are Chinese characters, we only remain the chinese token according to the training data.
Finally there are 5,201 characters left.
Similar to aforementioned works~\cite{39, 38}, we choose the metric of sentence level correction-f1 which could more strictly represent the classifier competence.

\textbf{Eurlex.}
This dataset~\cite{12} is for legal multi-label classification task. 
All EU laws are annotated by the EU Publications Office with multiple concepts from the eurovoc dictionary, a multilingual dictionary maintained by the Publications Office.
Given a document, the task is to predict its EuroVoc labels (concepts).
It contains 55,000/5,000/5,000 instances for training/validation/test.
There are 127 labels in this dataset and each sample has more than one label.
We use BERT as our base model and macro-f1/micro-f1 to evaluate the classifier performance.

\textbf{WOS-46985.}
We choose WOS-46985~\cite{13} as the dataset of multi-label classification task. 
Obviously, the total size of this dataset is 46,985. 
Following~\cite{43}, we split the dataset into training set, validation set and test set according to 6.4:1.6:2.
There are three types of labels: target label, parent label and child label in this dataset.
Similarly, we combined the parent label with the child label as the output classes following.
Finally, the number of output classes in this dataset is 141.
It is noted that there are two corresponding labels for each sample and we only select the largest two classes as the result in test.
We also use BERT as our backbone and macro-f1/micro-f1 as the evaluation metrics.

\textbf{WikiArt.}
WikiArt~\cite{56} is for image classification task, with the class of realism, art nouveau modern, analytical cubism and so on.
The following pre-processing is applied to each image: auto-orientation of pixel data with EXIF-orientation stripping and resize to 416 x 416.
We use ViT as our backbone and accuracy as the evaluation metric.

\textbf{Chest\_falsetto.}
We choose this dataset of the audio classification test.
The dataset contains 1,280 monophonic singing audio (.wav format) of chest and falsetto voices, with chest voice tagged as chest and falsetto voice tagged as falsetto.
We select Wav2vec2 as our backbone and accuracy as the evaluation metric.

\subsection{Baselines} \label{sec: baseline}
We implement 7 baselines including \textbf{Softmax} and its six variants.
We first compare our method with the temperature-based softmax (\textbf{T-Softmax}) which has hyper-parameter \textcolor{black}{$\tau$ to control the “softness” or “peakiness” of the output probability distribution}.
Then, for sparse based algorithms, we choose \textbf{Sparse-Softmax}~\cite{6}, \textbf{Sparsemax}~\cite{17} and \textbf{Entmax}~\cite{24} as our comparison methods.
From the perspective of the reachable training objective, we introduce \textbf{Label-Smoothing}~\cite{5} for comparison.
In addition, $\delta$ plays the role of margin in AS-Softmax.
Thus we compare AS-Softmax with the additive margin softmax (\textbf{AM-Softmax}~\cite{32}).
Most shared hyper-parameters refers to examples\footnote{\url{https://github.com/huggingface/transformers/tree/main/examples}}, while the specific hyper-parameters of each model are tuned on the validation set.
Details of these models and their hyper-parameter settings are shown below.

\textbf{T-Softmax.}
T-Softmax is a simple variant of introducing the temperature parameter $\tau$ into Softmax. 
\textcolor{black}{
The temperature parameter $\tau$ within the softmax function regulates the degree of smoothness or sharpness in the output probability distribution. 
It allows us to modulate the extent of randomness present in the function's output.
}
The prediction score of a class $p_i$ in T-Softmax is computed as follows:
\begin{align}
    p_i = \operatorname{T-Softmax}({o_i}) = \frac{e^{\frac{o_{i}}{\tau}}}{\sum_{j=1}^{n}e^{\frac{o_{j}}{\tau}}}. \label{eq:T-softmax}
\end{align}

\textbf{Label-Smoothing.}
The authors of Label-Smoothing~\cite{5} argue that these non-target classes need to be given a probability greater than zero so that the model can fully learn all classes' features.
Considering that the dataset is not completely correct, the original softmax loss function is even less applicable.
Label-Smoothing has a parameter $\epsilon$.
In the softmax loss function, the calculated weight of non-target classes is 0.
\textcolor{black}{To mitigate excessive confidence towards the ground truth, label smoothing assigns a weight of $\frac{\epsilon}{n-1}$ to non-target classes,} where $n$ represents the number of label classes.
The weight of the target class is $1-\epsilon$, where $\epsilon$ is a hyper-parameter of label smoothing.

\textbf{Sparse-Softmax.}
\textcolor{black}{
Sparse-Softmax~\cite{6} highlights that the outcomes derived from softmax activation lack sparsity, potentially causing overlearning.
}
It only preserves the features of the top $k$ classes, and the remaining features are directly assigned to 0 for mask.
Obviously, $k$ is a hyper-parameter.

\textbf{AM-Softmax.}
Inspired by algorithms such as large-margin, AM-Softmax~\cite{32} aims to minimize the intra-class variation.
Meanwhile, the loss function and back propagation process of large-margin softmax are too complicated due to the introduction of angle.
AM-Softmax not only simplifies the calculation process, but also speeds up the training speed.
Its loss function is as follows:
\begin{align}
    \mathcal{L} &= -\log \frac{e^{s\cdot(o_i - m)}}{e^{s\cdot (o_i - m)}+\sum^{n}_{j=1,j\neq i}e^{s\cdot {o_j}}}.
    \label{eq:am-softmax loss}
\end{align}
It has two parameters, $s$ and margin $m$ where $m$ is used to increase the margin between categories and $s$ has the effect of acceleration.
When $s$ or $m$ is too large, NAN may occur in the value of the loss function.
Therefore, $s$ and $m$ vary greatly with different datasets.
This is a process that needs to be adjusted slowly.

\textbf{Sparsemax.}
Sparsemax~\cite{17} returns sparse posterior distributions by assigning zero probability to its output variables with small scores.

\textbf{Entmax.}
Entmax~\cite{24} is a sparse family of probability mappings, generalizing softmax.
It has two versions including the sorting-based method and the bisection-based method.
Following experiment is conducted with the 1.5-entmax proposed by author which adopts the sorting-based method and $\alpha$ is equal to 1.5.

\begin{table*}[htbp]
    \centering
    \begin{tabular}{lc|c|c|c|c|c}
    \toprule
                                & \textbf{SST5}      & \textbf{Clinc\_oos} & \textbf{Conll2003} & \textbf{SIGHAN2015} & \textbf{Eurlex}    & \textbf{WOS-46985}  \\  \midrule \midrule
    T-Softmax ($\tau$)           & 5         & 0.5         & 0.5         & 0.5         & 1        & 0.5   \\
    Sparse-Softmax ($k$)           & 4         & 10         & 5         & 20         & 50        & 100   \\
    Entmax ($\alpha$)             & 1.5   & 1.5    & 1.5    & 1.5     & -    & -   \\
    Label-Smoothing ($\epsilon$) & 0.30      & 0.20       & 0.10      & 0.10       & -         & -     \\ 
    AM-Softmax ($s/m$)             & 10/0.30   & 8/0.35     & 5/0.30    & 1/0.10     & -    & -  \\ \midrule
    AS-Softmax ($\delta$/$r$)      & 0.30/0.00 & 0.30/0.15  & 0.35/0.25 & 0.30/0.00  & 0.05/0.00 & 0.35/0.00 \\ 
    AS-Speed ($\lambda$/$s$)       & 1.5/4     & 0.5/5      & 0.5/5     & 0.5/2      & 0.5/2     & 1.0/2  \\ \bottomrule
    \end{tabular}
    \caption{Parameter setting of algorithm. $s$ after AM-Softmax means scale. $s$ after AS-Speed means max accumulation steps.}
    \label{tab:parameter_results}
\end{table*}

\begin{table*}[htbp]
    \centering
    \begin{tabular}{lc|c|c|c|c|c}
    \toprule
                  & \textbf{SST5} & \textbf{Clinc\_oos} & \textbf{Conll2003} & \textbf{SIGHAN2015} & \textbf{Eurlex} & \textbf{WOS-46985} \\ \hline \midrule
    epoch         & 7    & 10         & 20        & 15         & 20     & 10    \\ 
    batch size    & 16   & 32         & 32        & 128        & 16     & 12   \\ 
    learning rate & 2e-5 & 2e-5       & 2e-5      & 5e-5       & 3e-5   & 3e-5  \\ \bottomrule
    \end{tabular}
    \caption{Information of universal hyper-parameters.}
    \label{tab:universal_parameter}
\end{table*}

\subsection{Settings}
Most experiments are conducted based on huggingface’s pytorch implementation of transformers\footnote{https://github.com/huggingface/transformers}.
The base pretrained models~\cite{44, 45} are initialized by BERT-base-cased and Roberta-wwm-ext respectively.
We utilize AdamW~\cite{46} as the optimizer.
Due to the little change of AS-Softmax to the original softmax code, it can be reproduced easily.
Our experiments are mainly carried out on RTX A5000 with a memory size of 24G except that the experiment related to SIGHAN2015 is conducted on V100 with 32G memory.

In our experiment, \textcolor{black}{the setting of $\delta$ depends on the difficulty of the task.
For most tasks, setting $\delta$ to 0.3 works well.
For simple tasks, a larger $\delta$ (around 0.35) is preferred.
For difficult tasks, a smaller $\delta$ (around 0.2) performs better.
In addition, the baseline performance of Clinc\_oos and Conll2003 is high, indicating their relatively lower task difficulty. 
Even if the model has not fully learned the category representations, it filters effective samples with the learning objective of AS-Softmax, leading to the discarding of some valid samples and affecting the overall performance.}
Thus, we additionally adopt special warm-up strategy on Clinc\_oos and Conll2003 by keeping $\delta$ equal to 1 in the first $r$ percent of training steps.
For AS-Speed algorithm, there are two more hyper-parameters: $\lambda$ and $s$.
$\lambda$ is a hyper-parameter which controls the accelerate magnitude while $s$ denotes maximum accumulated steps.
In addition, we implement Entmax, Sparsemax and AM-Softmax according to the source code\footnote{\url{https://github.com/deep-spin/entmax}}\footnote{\url{https://github.com/happynear/AMSoftmax}}.
We record the general and specific hyper-parameters respectively in Table~\ref{tab:parameter_results} and Table~\ref{tab:universal_parameter}.

\subsection{Main Results}
The performance and training time on experimental tasks are presented separately in Table~\ref{tab:main results} and Table~\ref{tab:running-time results}.
Overall, the proposed AS-Softmax and AS-Speed have achieved improvements in terms of classification performance and efficiency in almost all the tasks.

From Table~\ref{tab:main results}, we can observe that AS-Softmax consistently outperforms other methods.
For text multi-class classification tasks and text token classification tasks, we find that AS-Softmax could bring more performance enhancement in the case of difficult tasks than simple tasks.
It is worth noting that the f1 score of AS-Softmax on SIGHAN2015 is about 2 points higher than that of softmax.
In terms of text multi-label classification tasks, there is a slight difference between the training objective of AS-Softmax and the test goal.

In testing, we will choose the classes with scores greater than 0 as the prediction results.
However, in training period, AS-Softmax only encourages the probabilities of non-target classes to be lower than the minimum probabilities of target classes by a specific margin $\delta$.
Despite the impact of this gap, the results of AS-Softmax also surpass softmax results on such datasets.
More details will be analyzed in following sections.

Table~\ref{tab:running-time results} shows that all compared algorithms except AS-Speed perform fairly in terms of the training time.
Notably, when it comes to high-dimensional output classes (i.e. SIGHAN2015), it may cost more time for those methods which requires $O(nlogn)$ computation complexity, such as Entmax and Sparsemax.
Compared with softmax, AS-Speed could increase the training speed by about 1.2$\times$ on the premise of maintaining the classification accuracy.
To some extent, the acceleration effect depends on the number of classification categories.
AS-Speed has achieved greater speed improvement on SST5 and Conll2003 which have no more than 10 output classes.
Meanwhile, AS-Softmax also has more than 1.1$\times$ acceleration performance over softmax although there are 150+ output classes such as Clinc\_oos.
For SIGHAN2015 and WOS-46985, AS-Speed has no significant effect.
One important reason is that there are too many output classes and the proportion of masked samples is unstable, leading to the maximum accumulated steps can only be 2.

\begin{table*}[htbp]
  \centering
  \resizebox{0.95\textwidth}{!}{
    \begin{tabular}{lcc|cc|c|c|cc|cc}
    \toprule
          & \multicolumn{2}{c|}{\textbf{SST5}} & \multicolumn{2}{c|}{\textbf{Clinc\_oos}} & \textbf{Conll2003} & \textbf{SIGHAN2015} & \multicolumn{2}{c|}{\textbf{Eurlex}} & \multicolumn{2}{c}{\textbf{WOS-46985}} \\
          & \multicolumn{2}{c|}{accuracy} & \multicolumn{2}{c|}{accuracy} & \multirow{2}[1]{*}{f1} & \multirow{2}[1]{*}{f1} & \multirow{2}[1]{*}{marcro-f1} & \multirow{2}[1]{*}{micro-f1} & \multirow{2}[1]{*}{macro-f1} & \multirow{2}[1]{*}{micro-f1} \\
          & \multicolumn{1}{c}{BERT} & Roberta & \multicolumn{1}{c}{BERT} & Roberta &       &       &       &       &       &  \\
    \midrule \midrule
    Softmax & 51.90  & 55.57  & 88.60 & 90.76 & 90.56 & 70.80 & 46.87 & 68.73 & 80.40 & 86.00 \\
    T-Softmax & 53.12  & 56.56  & 88.65 & 89.91 & 90.63 & 70.80 & 46.87 & 68.73 & 80.76 & 86.36 \\
    Sparse-Softmax & 52.99  & 55.93  & 89.02 & 90.69 & 90.93 & 71.40 & 47.77 & \textbf{68.99} & 80.94 & 86.65 \\
    Sparsemax & 51.49  & 56.20  & 88.55 & \textbf{91.29} & 90.57 & 71.49 & -     & -     & -     & - \\
    Entmax & 51.90  & 55.07  & 88.33 & 89.84 & 90.62 & \textbf{72.81} & -     & -     & -     & - \\
    Label-Smoothing & 52.85  & 54.84  & 88.64 & 90.38 & 90.64 & 68.10 & -     & -     & -     & - \\
    AM-Softmax & 52.17  & 55.93  & 89.05 & 90.73 & 90.66 & 68.57 & -     & -     & -     & - \\
    \midrule
    AS-Softmax & \textbf{53.12 } & 57.29  & \textbf{89.07} & 90.96 & \textbf{91.03} & \textbf{72.81} & \textbf{48.30} & \textbf{68.99} & 80.94 & 86.39 \\
    AS-Speed & 52.53  & \textbf{57.87 } & 88.78 & 91.25 & 90.58 & 71.40 & 48.14 & 68.70 & \textbf{81.07} & \textbf{86.71} \\
    \bottomrule
    \end{tabular}
}
\caption
{\textcolor{black}{Comparison of experimental results on various text datasets with different baselines (``-'' means not available since their papers or codes do not provide the corresponding method for multi-label classification tasks).}
}
\label{tab:main results}
\end{table*}%

\begin{table*}[tp]
\centering
\small
\resizebox{0.7\textwidth}{!}{
    \begin{tabular}{lc|c|c|c|c|c}
    \toprule
                    & \textbf{SST5}         & \textbf{Clinc\_oos}   & \textbf{Conll2003}    & \textbf{SIGHAN2015}     & \textbf{Eurlex}        & \textbf{WOS-46985}   \\ \hline \midrule
    Softmax         & 288          & 533          & 625          & 10905          & 5944               & 5644     \\ 
    Sparse-Softmax  & -7\%          & +2\%          & -1\%          & -2\%          & -1\%          & -1\%        \\
    Sparsemax       & -2\%          & 0          &      0     & +8\%          & -             & -            \\ 
    Entmax       & 0          & 0          & -2\%          & +10\%          & -             & -            \\ 
    Label-Smoothing & 0          & +1\%          & -3\%          & -4\%          & -          & -          \\  
    AM-Softmax      & 0          & +1\%          & 0          & -3\%         &     -     &       -   \\\midrule 
    AS-Softmax      & -1\%          & -1\%          & -1\%          & -2\%          & -1\%          & -2\%         \\
    AS-Speed        & \textbf{-22\%} & \textbf{-15\%} & \textbf{-21\%} & \textbf{-7\%} & \textbf{-13\%} & \textbf{-5\%}   \\ \bottomrule
    \end{tabular}
}
\caption{Training time of different methods (in seconds). The training time of T-Softmax is not listed because it is the same as that of Softmax in theory.}
\label{tab:running-time results}
\end{table*}

\subsection{Result Analysis} \label{sec:consistent}

\begin{figure*}[htbp]
    \centering
    \includegraphics[width=1\linewidth]{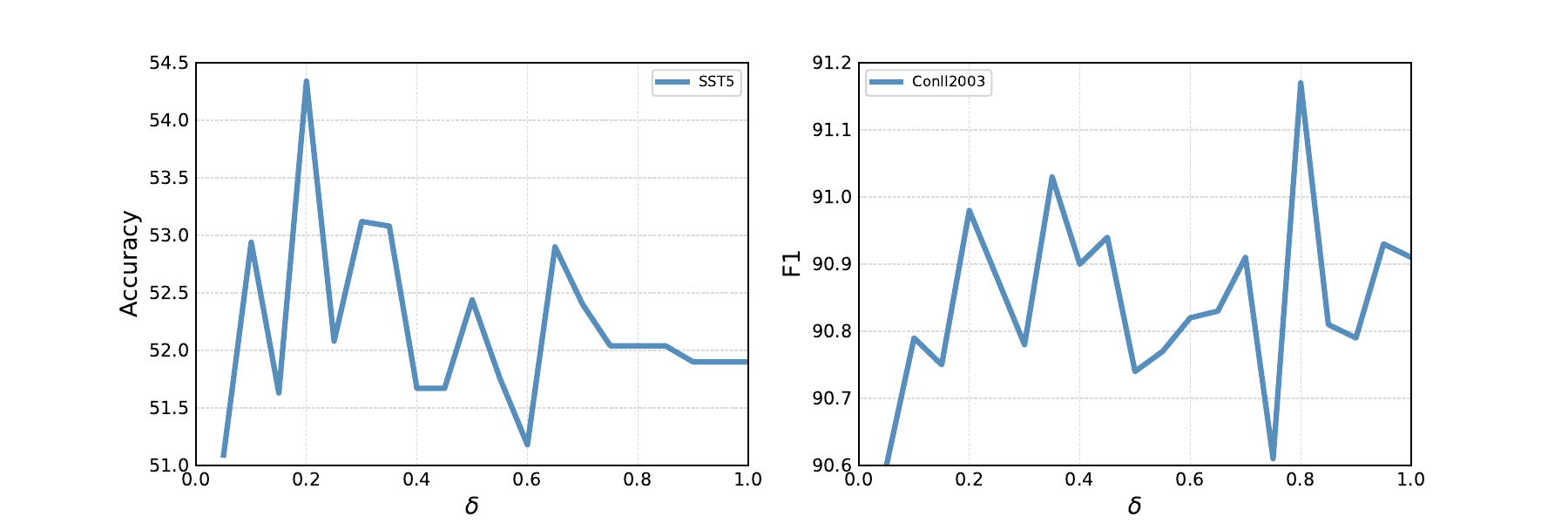}
    \caption{\textcolor{black}{Accuracy/F1 metrics on SST5 and Conll2003 with different $\delta$.}}
    \label{fig:different delta}
\end{figure*}
\paragraph{Impact of $\delta$}
The $\delta$ can significantly affect the accuracy of the classifier.
If $\delta$ is too large, AS-Softmax would not mask such easy samples, making the model to overlearn a lot of useless information.
In contrast, AS-Softmax would cast away some potential useful knowledge when $\delta$ is rather small, which could result in insufficient learning.
In our experiments, we attempt different values of $\delta$ ranging from \textcolor{black}{0} to \textcolor{black}{1} and choose the final value which achieves the highest classification performance on the development set. 
We investigate the influence of $\delta$ in the case of relative easy and hard tasks.
As Figure~\ref{fig:different delta} shown, the predictions on SST5 represents a difficult situation while the result on Conll2003 denotes a simple case.

\textcolor{black}{For the SST5 dataset, the performance of the original softmax method is around 51.90. 
This indicates that the task is quite challenging, and the ability of classifier to distinguish between confusing categories is weak. 
If we set the delta value relatively high, e.g. 0.9, the classifier is highly likely to fail to meet this learning objective. No samples will be discarded, causing AS-Softmax to degrade into Softmax, offering no significant benefit to training. 
Therefore, a relatively small value, such as 0.2, is able to ultimately achieve decent performance.
However, for the Conll2003 dataset, the original softmax method already achieves around 90.56 in performance. 
This suggests that the task is not highly challenging, and the classifier demonstrates strong capabilities in distinguishing between closely related categories. 
Consequently, setting delta to a larger value might yield a favorable effect.}
According to such experiments, we suggest setting $\delta$ to a moderate value 0.3 and we believe that it could be better to set $\delta$ to a relative larger value for simple tasks and set $\delta$ to a smaller value for difficult tasks.

\begin{figure*}[htbp]
    \centering
    \includegraphics[width=0.8\linewidth]{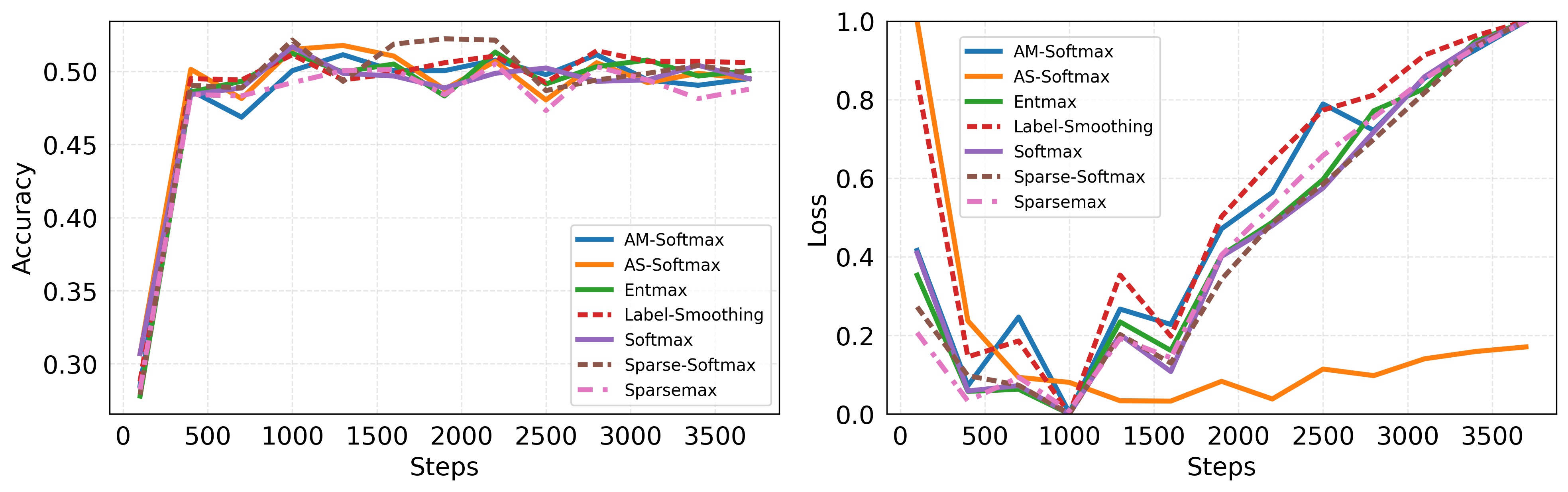}
    \caption{Comparison of development performance on SST5.}
    \label{fig:loss curves}
\end{figure*}

\paragraph{Correlation between Loss and Classification Performance}
To verify that our training objective can alleviate the train-test goal mismatching problem,
we \textcolor{black}{employ} the Pearson correlation coefficient which measures the relationship between the loss and the classification ability in validation.
The correlation coefficient is close to -1 or 0 indicating the loss is linear or has no significant correlation with the classification performance.
If the correlation coefficient is close to 1, it demonstrates that the learning direction of the model is obviously opposite.
In theory, the final classification performance is inversely correlated with the loss.
\textcolor{black}{Thus, the ideal correlation coefficient is -1, and a closer approximation to -1 indicates superior model performance.}
Table~\ref{tab:pearsonr results} illustrates that the correlation between loss of all methods and their classification performance is linear on datasets with high identical accuracy such as Conll2003 and SIGHAN2015.
However, when it comes to hard datasets (i.e. SST5 and Eurlex), the correlation coefficients of compared baselines are not good enough and some correlations could be even the opposite.
By comparison, the loss of AS-Softmax and its classification performance is highly relevant across whole experimental datasets, which verified the design of AS-Softmax training objective.

To intensively explore the correlation between the loss and classification performance, we record the loss and accuracy on development set of SST5.
The left/right subfigure of Figure~\ref{fig:loss curves} represents its classification performance and normalized loss, respectively.
It can be seen that other models tend to overfit as their accuracy is still fluctuating while the loss has turned to increase, after 1000+ training steps.
By contrast, the loss of AS-Softmax does not increase significantly at the end of training process, which also reveals that it could moderate the overfitting problem.

To verify the ``easy'' samples \textcolor{black}{are not selected by chance}, we construct the datasets with hard samples left at the end of the AS-Softmax training period on SST5 (approximately 11$\%$ of all) and the same number of random selected samples. 
Then, we train the classifier with softmax and the result is shown in the Table~\ref{tab:hard samples left}. 
The performance illustrates that these hard samples are more difficult than the random selected samples for the model to learn.

\begin{table*}[htbp]
\centering
\small
\resizebox{0.8\textwidth}{!}{
\begin{tabular}{lcc|cc|cc}
\toprule
                  & \multicolumn{1}{c|}{\textbf{SST5}}            & \textbf{Clinc\_oos}      & \multicolumn{1}{c|}{\textbf{Conll2003}}       & \textbf{SIGHAN2015}      & \multicolumn{1}{c|}{\textbf{Eurlex}}         & \textbf{WOS-46985}        \\ \hline \midrule
Softmax           & \multicolumn{1}{c|}{0.038}           & -0.943          & \multicolumn{1}{c|}{-0.975}          & -0.947          & \multicolumn{1}{c|}{-0.337}         & -0.902          \\ 
Sparse-Softmax    & \multicolumn{1}{c|}{0.088}           & -0.914          & \multicolumn{1}{c|}{-0.981}          & -0.907 & \multicolumn{1}{c|}{-0.114}         & -0.901          \\ 
Sparsemax         & \multicolumn{1}{c|}{0.050}            & -0.976 & \multicolumn{1}{c|}{-0.968}          & -0.681          & \multicolumn{1}{c|}{-}              & -               \\ 
Entmax         & \multicolumn{1}{c|}{0.080}            & \textbf{-0.986} & \multicolumn{1}{c|}{-0.983}          & -0.617          & \multicolumn{1}{c|}{-}              & -               \\ 
Label-Smoothing   & \multicolumn{1}{c|}{-0.087}          & -0.952          & \multicolumn{1}{c|}{\textbf{-0.998}} & \textbf{-0.989}          & \multicolumn{1}{c|}{-}              & -               \\ 
AM-Softmax     & \multicolumn{1}{c|}{0.057}          & -0.915          & \multicolumn{1}{c|}{-0.974}          & -0.982          & \multicolumn{1}{c|}{-}          &   -       \\ \midrule
AS-Softmax        & \multicolumn{1}{c|}{\textbf{-0.952}} & -0.949          & \multicolumn{1}{c|}{-0.996}          & -0.986          & \multicolumn{1}{c|}{\textbf{-0.909}} & \textbf{-0.904} \\ \bottomrule
\end{tabular}
}
\caption{Results of Pearson correlation coefficient between loss and accuracy.}
\label{tab:pearsonr results}
\end{table*}

\begin{figure*}[htbp]
    \centering
    \includegraphics[width=0.8\linewidth]{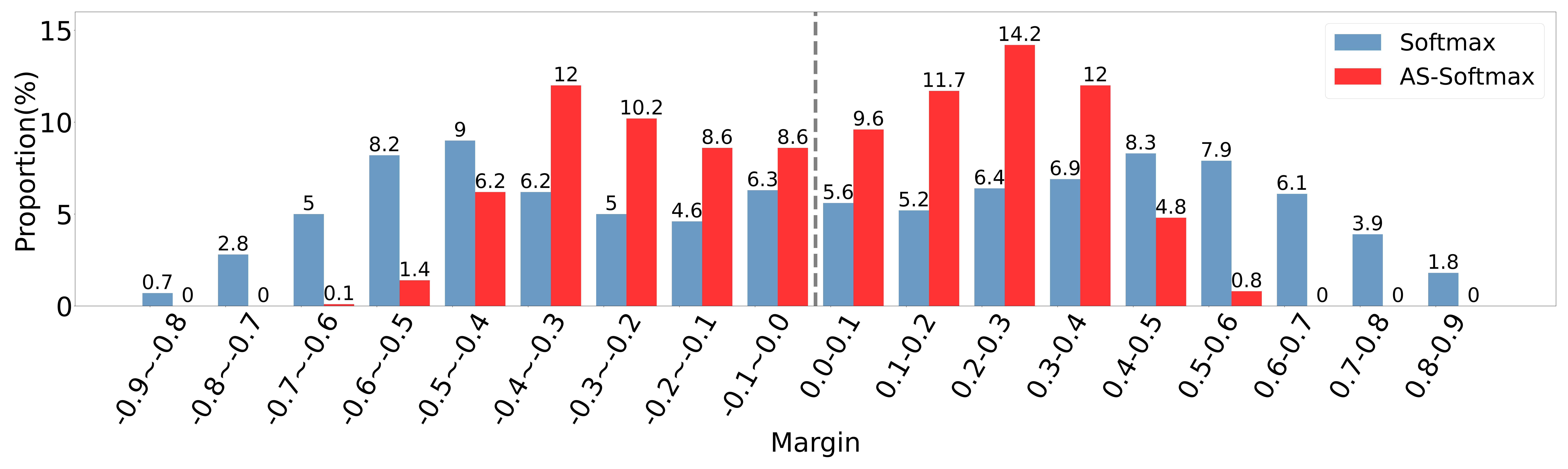}
    \caption{Distribution diagram of $p_{margin}$ on the SST5 test set with $\delta = 0.3$. 
    We add a dotted line to stand for $p_{margin} = 0$, which splits the  wrongly (left) and correctly (right) predicted cases.}
    \label{fig:label_neg1}
\end{figure*}
\begin{table}[htbp]
    \centering
    \resizebox{0.2\textwidth}{!}{
        \begin{tabular}{lc}
        \toprule
                                    & \textbf{SST5}        \\  \midrule 
        Random samples           & 45.52       \\
        Hard samples             & 38.64    \\ \bottomrule
        \end{tabular}
    }
    \caption{\textcolor{black}{Accuracy} of standard Softmax on random sample dataset and hard sample dataset.}
    \label{tab:hard samples left}
\end{table}

\begin{table*}[ht]
  \centering
  \small
  {
    \begin{tabular}{lc|c|c|c|c|c|c|c}
    \toprule
\textbf{}                & \textbf{MNLI:3} & \textbf{MRPC:2} & \textbf{QNLI:3} & \textbf{QQP:2} & \textbf{RTE:3} & \textbf{SST-2:2} & \textbf{CoLA:2} & \textbf{Avg}   \\ \midrule\midrule
Softmax         & 84.18           & 89.41           & 90.49           & 91.00          & 68.23          & 92.54            & 62.44           & 82.61          \\
\textcolor{black}{T-Softmax}      & \textcolor{black}{83.75}           & \textcolor{black}{89.77}           & \textcolor{black}{90.54}           & \textcolor{black}{90.99}          & \textcolor{black}{67.51}          & \textcolor{black}{92.43}            & \textcolor{black}{62.10}           & \textcolor{black}{82.44}          \\
\textcolor{black}{Sparsemax}       & \textcolor{black}{83.59}           & \textcolor{black}{89.50}           & \textcolor{black}{90.68}           & \textbf{\textcolor{black}{91.12}} & \textcolor{black}{67.51}          & \textcolor{black}{92.32}            & \textcolor{black}{61.83}           & \textcolor{black}{82.36}          \\
\textcolor{black}{Entmax}          & \textcolor{black}{83.98}           & \textcolor{black}{88.70}           & \textcolor{black}{90.72}           & \textcolor{black}{91.00}          & \textcolor{black}{67.51}          & \textcolor{black}{92.66}            & \textcolor{black}{62.63}           & \textcolor{black}{82.46}          \\
\textcolor{black}{Label-Smoothing} & \textcolor{black}{83.53}           & \textcolor{black}{88.81}           & \textcolor{black}{90.85}           & \textcolor{black}{91.03}          & \textcolor{black}{68.59}          & \textcolor{black}{92.78}            & \textcolor{black}{63.33}           & \textcolor{black}{82.70}          \\
\textcolor{black}{AM-Softmax}      & \textcolor{black}{81.88}           & \textcolor{black}{90.38}           & \textcolor{black}{90.76}           & \textcolor{black}{89.29}          & \textcolor{black}{68.23}          & \textcolor{black}{92.78}            & \textcolor{black}{63.48}           & \textcolor{black}{82.40}          \\
\midrule
AS-Softmax      & \textbf{84.19}  & \textbf{90.72}  & \textbf{91.12}  & 90.95          & \textbf{69.31} & \textbf{93.23}   & \textbf{64.55}  & \textbf{83.44} \\
\textcolor{black}{AS-Speed}        & \textcolor{black}{83.73}           & \textcolor{black}{90.43}           & \textcolor{black}{90.99}           & \textcolor{black}{90.67}          & \textcolor{black}{68.95}          & \textcolor{black}{92.78}            & \textcolor{black}{62.91}           & \textcolor{black}{82.92}         \\ \bottomrule
\end{tabular}%
    }
    \caption{
    Result of GLUE benchmark (better results are in \textbf{bold}), with the number of output classes attached behind the task name.}
  \label{tab:glue_result}
\end{table*}%

\begin{table*}[ht]
  \centering\textcolor{black}{
  \small
  {
    \begin{tabular}{l|ccccc|c}
    \toprule
     & \textbf{\textcolor{black}{\# Very Negative}} & \textbf{\textcolor{black}{\# Negative}} & \textbf{\textcolor{black}{\# Neutral}} & \textbf{\textcolor{black}{\# Positive}} & \textbf{\textcolor{black}{\# Very Positive}} & \textbf{\textcolor{black}{Accuracy (Softmax/AS-Softmax)}} \\ \midrule\midrule
\textcolor{black}{SST5}           & \textcolor{black}{1092}             & \textcolor{black}{2218}             & \textcolor{black}{1624}             & \textcolor{black}{2322}             & \textcolor{black}{1288}             & \textcolor{black}{51.90 / \textbf{53.12}}                           \\ \midrule
\multicolumn{7}{c}{\textbf{\textcolor{black}{Setting 1}}} \\ \midrule
\textcolor{black}{{After sampling}} & \textcolor{black}{200}              & \textcolor{black}{2218}             & \textcolor{black}{200}              & \textcolor{black}{2322}             & \textcolor{black}{200}              & \textcolor{black}{49.19 / \textbf{49.91}}                           \\
\textcolor{black}{{After sampling}} & \textcolor{black}{300}              & \textcolor{black}{2218}             & \textcolor{black}{300}              & \textcolor{black}{2322}             & \textcolor{black}{300}              & \textcolor{black}{51.22 / \textbf{52.22}}                           \\
\textcolor{black}{{After sampling}} & \textcolor{black}{400}              & \textcolor{black}{2218}             & \textcolor{black}{400}              & \textcolor{black}{2322}             & \textcolor{black}{400}              & \textcolor{black}{49.95 / \textbf{51.36}}                           \\
\textcolor{black}{{After sampling}} & \textcolor{black}{500}              & \textcolor{black}{2218}             & \textcolor{black}{500}              & \textcolor{black}{2322}             & \textcolor{black}{500}              & \textcolor{black}{51.18 / \textbf{52.90}}  \\ \midrule
\multicolumn{7}{c}{\textbf{\textcolor{black}{Setting 2}}}                                                                                                                        \\ \midrule
\textcolor{black}{{After sampling}} & \textcolor{black}{100}              & \textcolor{black}{500}              & \textcolor{black}{100}              & \textcolor{black}{500}              & \textcolor{black}{100}              & \textcolor{black}{45.88 / \textbf{48.69}}                           \\
\textcolor{black}{{After sampling}} & \textcolor{black}{100}              & \textcolor{black}{800}              & \textcolor{black}{100}              & \textcolor{black}{800}              & \textcolor{black}{100}              & \textcolor{black}{47.65 / \textbf{49.10}}                           \\
\textcolor{black}{{After sampling}} & \textcolor{black}{100}              & \textcolor{black}{1500}             & \textcolor{black}{100}              & \textcolor{black}{1500}             & \textcolor{black}{100}              & \textcolor{black}{47.06 / \textbf{48.51}}                          
\\
\bottomrule                       
\end{tabular}%
    }
    \caption{\textcolor{black}{
    Accuracy results on the SST5 dataset with different imbalance settings. SST5 contains five classes including very negative, negative, neural, positive and very positive.
}}
  \label{tab:imbalance_result}}
\end{table*}%

\paragraph{Probability Margin of Positive and Negative Classes}
The AS-Softmax training goal is calculated according to target probability $p_t$ and non-target probability $p_{i \neq t}$.
To check this objective, we design a metric $p_{margin} = p_t - p_{neg}^{max}$, where $p_{neg}^{max}$ indicates the largest probability in non-target classes.
$p_{margin}$ is greater or less than 0 denoting that the classification result is correct or wrong.
Then we draw the distribution diagram of $p_{margin}$ on SST5 in Figure~\ref{fig:label_neg1}, where $\delta$ in AS-Softmax is set to 0.3.
Generally speaking, Softmax result is widely distributed while the distribution of AS-Softmax result is very concentrated.
On one hand, when $p_{margin}$ is less than 0, AS-Softmax pushes $p_{margin}$ close to 0.
It indicates that even if the true label is not identified correctly, its top k candidates in the prediction result still have some reference value.
By comparison, there are many strong negative classes in the softmax result, which makes the result completely unavailable.
On the other hand, when $p_{margin}$ is greater than 0, the overall distribution of $p_{margin}$ is squeezed to be lower than or close to $\delta$.
It suggests that AS-Softmax could prevent the model from being optimized towards overlearning easy samples and the model could pay more attention to those hard samples.
To conclude, the probability margin distribution is in accordance with the proposed training objective which encourages the target class to exceed other non-target classes by a specific margin $\delta$.

\begin{figure}[htbp]
    \centering
    \includegraphics[width=.8\linewidth]{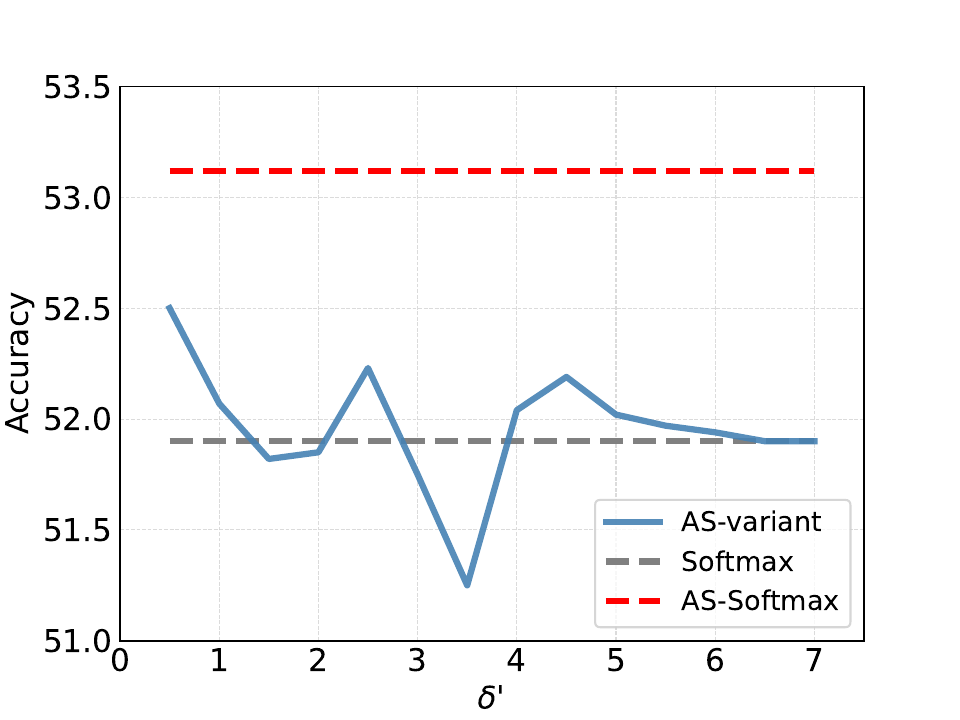}
    \caption{Accuracy of AS-variant with different $\delta'$ on SST5 dataset.}
    \label{fig:AS-Softmax variant}
\end{figure}
\paragraph{AS-variant}

To compare the difference between training objective of Eq.~\ref{eq:am-softmax loss} and Eq.~\ref{eq:log_delta}, we conduct experiment using them respectively, where the latter we called AS-variant.
Similar to AS-Softmax, we applying $\delta'$ to the final output $o$ of the neural classifier where $\delta'$ is a margin used to widen the gap between the scores of target categories and others.
One obvious difference is that $\delta$ in AS-Softmax is in $[0,1]$ while $\delta'$ in AS-variant has no upper bound because $o$ can take any value.
It makes it hard to find the optimal value.

The comparison between AS-variant and AS-Softmax on SST5 is shown in Figure~\ref{fig:AS-Softmax variant}.
From this figure, the gray dotted line represents the prediction results of original softmax while the experimental results of AS-Softmax are indicated by the red dotted line.
The blue one represents that the results of AS-variant change with $\delta'$.
We take the value of $\delta'$ from 0 to 7.
As can be seen from this figure, when $\delta'$ is greater than 6, AS-variant is equivalent to softmax.
It can be seen that the prediction results fluctuate up and down around the baseline.
Therefore, we can not determine where to obtain the optimal $\delta'$ under this algorithm as easily as AS-Softmax.
Moreover, the results are not as good as those of AS-Softmax.

\begin{table}
    \centering
    \textcolor{black}{
    \begin{tabular}{lccc}
    \toprule
         & \textbf{SST5} & \textbf{Clinc\_oos} & \textbf{Conll2003} \\
    \midrule
    Softmax     & 51.90 & 88.60 & 90.56\\
    Sparsemax     & 51.49 & 66.55 & 90.57\\
    Sparsemax+acceleration     & 51.49 & 89.62 &90.71 \\
    \bottomrule
    \end{tabular}
    \caption{\textcolor{black}{Accuracy of the standard softmax and Sparsemax with or without acceleration method.}}
    \label{tab:performance of as-speed on other baseline}}
\end{table}

\begin{table}
    \centering
    \textcolor{black}{
    \begin{tabular}{lccc}
    \toprule
         & \textbf{SST5} & \textbf{Clinc\_oos} & \textbf{Conll2003} \\
    \midrule
    Softmax     & 288 & 533 & 625\\
    Sparsemax     & -2\% & 0 & 0\\
    Sparsemax+acceleration     & -10\% & -13\% & -16\% \\
    \bottomrule
    \end{tabular}
    \caption{\textcolor{black}{Time consumption of the standard softmax and Sparsemax with or without acceleration method.}}
    \label{tab:time of as-speed on other baseline}}
\end{table}

\paragraph{The Effect of Accelerating Strategy}
AS-Speed accelerating strategy works by reducing the number of optimizing steps.
To explore the benefits of the adaptive gradient accumulation strategy, we attempt to use the same idea to accelerate other methods.
Our AS-Speed strategy takes into account that, as the training progresses, the training objective of AS-Aoftmax will result in a lower effective sample size per batch compared to softmax. 
Among other variants of softmax, Sparsemax also reduces the effective sample size within a batch. 
Hence, we validated the performance of our adaptive gradient accumulation strategy when applied to Sparsemax on three dataset, as shown in the Table~\ref{tab:performance of as-speed on other baseline} and Table~\ref{tab:time of as-speed on other baseline}. 
Additionally, the advantage of AS-Speed lies in accelerating the training process without affecting the model's performance.

\paragraph {Results on GLUE Datasets}
To verify the effectiveness of AS-Softmax on general text classification tasks, we conduct additional experiment on some classification tasks of GLUE benchmark~\cite{47}, including MNLI, MRPC, QNLI, QQP, RTE, SST-2 and CoLA.
The shared hyper-parameters are set following~\cite{45} while the margin $\delta$ in AS-Softmax is adjusted in [0.05, 0.35] for searching.
\textcolor{black}{
Notably, Sparse-Softmax involves retaining the top-k candidate classes during model training. 
However, for the seven tasks in GLUE, which only consist of 2-class and 3-class classification, making the Sparse-Softmax method is not suitable.
}
Table~\ref{tab:glue_result} presents the experimental result.
It can be seen that the result of AS-Softmax also surpasses that of Softmax \textcolor{black}{and other baselines}.

\textcolor{black}{\paragraph{Results on imbalanced dataset}
We employed two settings to adjust the class distribution in the dataset, exploring the performance of AS-Softmax when the dataset is imbalanced. 
Taking the SST5 dataset as an example, we considered the category with higher proportion in the original dataset as the major classes and the others as minor classes. 
On one hand, we fixed the quantity of the major class and reduced the quantity of minor classes to around 1/10 of its original size.
On the other hand, we kept the quantity of minor classes constant and proportionally increased the quantity of the major class to induce data imbalance. 
Specifically, in the setting 1, we retain all samples from major classes (i.e. class negative and class positive), while subsampling an equal number of 200, 300, 400, and 500 samples from minor classes to assess the performance of AS-Softmax; in the setting 2, we fix the number of minor classes to 100, and subsample samples from major classes in proportions of 1:5, 1:8 and 1:15, respectively.
As illustrated in Table~\ref{tab:imbalance_result}, the result indicated that even in the presence of imbalanced class distribution in the dataset, AS-Softmax still provided a significant improvement over the original Softmax.
}

\begin{table}[htbp]
  \centering
  \textcolor{black}{
    \begin{tabular}{lcc|cc}
    \toprule
          & \multicolumn{2}{c|}{\textbf{Accuracy}} & \multicolumn{2}{c}{\textbf{Time Consumption}} \\
          & \textbf{Image} & \textbf{Audio} & \textbf{Image} & \textbf{Audio} \\
    \midrule
    Softmax & 46.74 & 67.97 & 996   & 199 \\
    T-Softmax & 49.76 & 69.53 & -     & - \\
    Sparse-Softmax & 50.87 & 72.66 & 1\%   & 2\% \\
    Sparsemax & 48.33 & 73.44 & 1\%   & 0 \\
    Entmax & 51.19 & 75.78 & 2\%   & 5\% \\
    Label-Smoothing & 49.92 & 72.66 & -1\%  & 5\% \\
    AM-Softmax & 49.92 & 74.22 & 1\%   & 6\% \\
    \midrule
    AS-Softmax & \textbf{51.19} & \textbf{76.56} & -1\%  & 4\% \\
    AS-Speed & 50.40  & 73.44 & \textbf{-10\%} & \textbf{-13\%} \\
    \bottomrule
    \end{tabular}%%
  \caption{\textcolor{black}{Accuracy and time consumption of the image and audio modal dataset.}}
  \label{tab:image_audio_result}}
\end{table}%

\textcolor{black}{\paragraph{Discussion about the interpretability}
To further explore the interpretability of AS-Softmax, we conducted additional experiments using other backbone with the learning objective of AS-Softmax and examined whether the conclusions aligned with previous findings. 
Here, we conducted supplementary experiments utilizing Roberta as the base model. The experimental results are presented in Table~\ref{tab:main results}.
It's observable that due to Roberta's superior performance compared to BERT, the Softmax method achieves higher results while that of BERT is 51.90. 
Additionally, with the assistance of AS-Softmax, the performance of model experiences further enhancement, especially on the difficult task of the SST5 dataset. 
This finding remains consistent with the results when BERT was used as the baseline, further validating the interpretability of our approach.
}

\textcolor{black}{\paragraph{Results on Other Modal Datasets}
We choose two datasets, one from the image domain and the other from the audio domain, to validate the performance of AS-Softmax. 
The visual models trained by WikiArt~\footnote{\url{https://huggingface.co/datasets/keremberke/painting-style-classification}} can be used to identify art from 27 different artistic movement categories.
Dataset Chest\_falsetto~\footnote{\url{https://huggingface.co/datasets/ccmusic-database/chest_falsetto}}contains 1,280 monophonic singing audio (.wav format) of chest and falsetto voices, with chest voice tagged as chest and falsetto voice tagged as falsetto.
The results in the Table~\ref{tab:image_audio_result} indicate a 4.5\% improvement in the image classification task and an 8.6\% improvement in the audio classification task with AS-Softmax. 
Additionally, the AS-Speed algorithm significantly reduces model training time, approximately by 10\% and 13\%, while maintaining similar improvements in performance.
}

\section{Conclusion}
In this paper, we propose a simple yet effective softmax variant, namely adaptive sparse softmax (AS-Softmax), which discards easy training samples at the aim of a more reasonable and test-matching learning objective.
We further develop an adaptive gradient accumulation strategy based on the masked sample ratio to accelerate the training process of AS-Softmax.
Experimental results on 6 text classification datasets show that AS-Softmax consistently surpasses the original softmax and its variants.
The limitation of AS-Softmax lies in the need for multiple trials to find the most suitable hyperparameters due to the introduction of additional parameters like $\delta$ and ratio $r$.
We believe our work can be extended in many aspects.
\textcolor{black}{For example}, instead of the current fixed value of $\delta$, we plan to investigate an improvement strategy which decides $\delta$ adaptively.

\end{document}